\documentclass[review]{elsarticle}

\usepackage{hyperref}
\usepackage{amsmath}
\usepackage{amssymb}
\usepackage{booktabs}
\usepackage{algorithm,algorithmic}
\usepackage{color}
\usepackage{subfig}
\biboptions{sort&compress} 
\journal{Journal of \LaTeX\ Templates}

\begin{document}
\begin{frontmatter}

\title{Content and Context Features for Scene Image Representation}




\author[Affil1]{Chiranjibi Sitaula \corref{cor1}}
\cortext[cor1]{Corresponding Author}
\ead{csitaul@deakin.edu.au}
\author[Affil1]{Sunil Aryal}
\author[Affil1]{Yong Xiang}
\author[Affil2]{Anish Basnet}
\author[Affil1]{Xuequan Lu}
\address[Affil1]{School of Information Technology, Deakin University, Geelong, Victoria 3216, Australia}
\address[Affil2]{Ambition College, Kathmandu, Nepal}

\begin{abstract}
Existing research in scene image classification has focused on either content features (e.g., visual information) or context features (e.g., annotations). As they capture different information about images which can be complementary and useful to discriminate images of different classes, we believe the fusion of them will improve classification results. In this paper, we propose new techniques to compute content features and context features, and then fuse them together. For content features, we design multi-scale deep features based on background and foreground information in images. For context features, we use annotations of similar images available in the web to design a codebook (filter words). Our experiments on three widely used benchmark scene datasets using Support Vector Machine (SVM) classifier reveal that our proposed context and content features produce better results than existing context and content features, respectively. The fusion of the proposed two types of features significantly outperform numerous state-of-the-art features.

\end{abstract}

\begin{keyword}
Image classification\sep Context features\sep Content features\sep Feature extraction\sep Machine learning \sep Image processing.
\end{keyword}

\end{frontmatter}


\section{Introduction}
Image representation in a machine readable form is an emerging field due to the wide use of camera technology in our daily life \cite{sitaula2019indoor}. For automatic analysis, images are often represented as a set of features. Most prior works in image feature extraction are focused on the content of images \cite{zeglazi_sift_2016, oliva2005gist, oliva_modeling_2001, dalal2005histograms,lazebnik2006beyond}. Features are extracted based on the visual information of images such as pixels, colors, objects, scenes, and so on. Compared to the existing traditional vision-based content features \cite{zeglazi_sift_2016, oliva2005gist, oliva_modeling_2001, dalal2005histograms,lazebnik2006beyond, wu_centrist:_2011}, content features extracted from the pre-trained deep learning methods \cite{he2016deep, zhou2016places, zhang2017image,tang_g-ms2f:_2017,8085139,guo2016bag} are found to be more effective. 

\begin{figure}[b]
\begin{center}
 \subfloat[]{\includegraphics[width=0.45\textwidth, height=30mm,keepaspectratio]{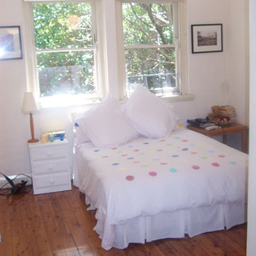}}
\hspace{10pt}
 \subfloat[]{\includegraphics[width=0.50\textwidth, height=30mm,keepaspectratio]{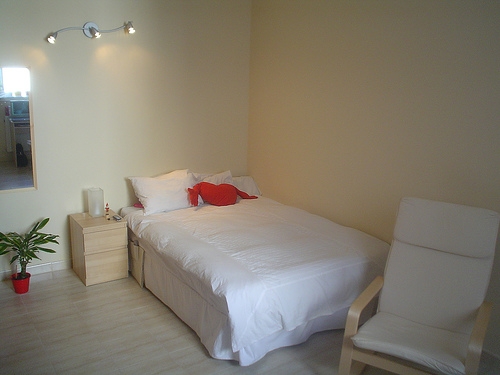}}
   \caption{The images of bed room (a) and hospital room (b) look similar on the basis of content (background and foreground) information. }
  \label{fig:1}
 \end{center}
  \end{figure}

Though these content features are shown to work reasonably well in many image processing tasks \cite{sitaula2019unsupervised,sitaula2020attention,sitaula2020fusion}, the information they capture may be insufficient to discriminate complex and ambiguous images with inter-class similarities and intra-class dissimilarities such as scene images. Fig.~\ref{fig:1} shows two images which look very similar but they belong to two different categories (hospital room and bedroom). In such a case, the contextual information of images can provide rich discriminating information. Contextual information about an image can be usually obtained from its annotations or descriptions. Though it is impossible to have descriptions or annotations for all images, such information can be searched from the web for certain types of images like objects and scenes. A very few studies in the literature considered the contextual information in scene image classification \cite{zhang2017image,wang2019task}. They extracted tag-based features from description/annotations of similar images. 

Despite that tag-based features produce better results than traditional vision-based content features and are comparable to some deep content features \cite{zhang2017image,wang2019task}, they still have some limitations. In general, they have a large feature size and low classification accuracy.
They use task-generic filter banks, which may miss the relevant tags in most cases. Furthermore, they also suffer from the repetition of semantically similar words such as plane and planes. These issues deteriorate the classification performance of scene images. In this paper, we introduce a new technique to extract tag features based on the pre-trained word embedding vectors \cite{mikolov2013efficient,pennington2014glove,bojanowski2017enriching}, which are used to define the associations between words/tags semantically.

For content-based features, we adopt features extracted from pre-trained deep learning models. In most state-of-the-art methods, deep features are extracted
from the intermediate layers of VGG16 model \cite{simonyan2014very}, GoogleNet \cite{szegedy2015going}, ResNet-152 \cite{he2016deep}, etc. 
Nevertheless, normally these features from such pre-trained models are extracted based on the fixed input image size of $224 \times 224$ pixels. { This is because pre-trained VGG-16 models have been trained with images of $224 \times 224$ pixels size during training}. 
Existing methods have a limited capability to deal with varied sizes of images and may not yield stable performance in the case of multi-scale (or also called multi-size) images.
{
Also, existing deep content features based on ImageNet and Places pre-trained deep learning models impart foreground and background information, respectively.} However, both information are equally important for scene image representation. 
As such, the multi-scale deep features based on foreground and background information are needed to achieve good classification performance.

In literature, either content-based or context-based features have been mostly used for image classification. Since these two sets of features capture different types of information about images, which may be complementary in many cases, we suppose that their fusion will boost the classification accuracy. Therefore, we propose to combine the two sets of features to produce Content and Context Features (abbreviated as $CCF$).

The \textbf{main contributions} of this paper are summarized as follows.
\begin{enumerate}
    \item We propose a {\bf novel approach} to extract context features, {which is tag-based features}, of an image from tags in annotations/descriptions of its top $k$ similar images available on the web. 
    For this, we first design a novel filter bank (codebook) with a size significantly smaller than the number of raw tags by eliminating redundant and outlier tags. {Note that before achieving our codebook, we also devise a novel algorithm to perform the stemming of tokens during pre-processing step.}
    Last, the {\bf t}ag-based context {\bf f}eatures ($TF$) of the image are extracted as the histogram based on the codebook, leading to features with a noticeably smaller size compared to existing tag-based context features. 
    \item For content features, we design multi-scale deep features based on background and foreground information in images. Our {\bf d}eep {\bf f}eatures are multi-scale 
    and have a smaller size compared to existing deep content features.
    \item We propose to fuse {\bf c}ontext and {\bf c}ontent {\bf f}eatures ($CCF$), to enable the ability of capturing different information in discriminating scene images of different classes. 
    \item To validate the proposed tag-based (context) features ($TF$), deep (content) features ($DF$) and the fusion of the two types of features ($CCF=TF+DF$), we evaluate their performances against state-of-the-art features in the scene image classification task using Support Vector Machine (SVM) on three commonly used scene image datasets (MIT-67~\cite{quattoni_recognizing_2009}, Scene-15~\cite{fei-fei_bayesian_2005} and Event-8~\cite{li2007and}). The results show that our proposed $TF$ and $DF$ produce better results than many existing context and content features, respectively and the combined $CCF$ produce significantly better results than numerous existing tag-based and deep learning-based features. 
\end{enumerate}

The remainder of the paper is organized as follows. Section \ref{lit:review} reviews related works in terms of content and context features extraction. Section \ref{prop:method} discusses the proposed methods to extract context-based, content-based and fused features. Section \ref{experiments} details the implementation and results of our experiments in scene image classification. 
Finally, we conclude the paper and point out potential future work in the last section.

\section{Related Works}
\label{lit:review}
Depending on the feature extraction source, there are basically two different groups of scene image features: content features and context features. The two types of features are explained separately in the following sub-sections.

\subsection{Content features}
Most of the scene image features are based on the content of the images ~\cite{zeglazi_sift_2016,oliva2005gist,oliva_modeling_2001, dalal2005histograms,lazebnik2006beyond,wu_centrist:_2011,xiao_mcentrist:_2014, margolin2014otc,quattoni_recognizing_2009,zhu_large_2010,li2010object,parizi2012reconfigurable,juneja2013blocks,lin_learning_2014,shenghuagao2010local,perronnin2010improving,gong_multi-scale_2014, kuzborskij2016naive,he2016deep, zhou2016places,zhang2017image,tang_g-ms2f:_2017,8085139,guo2016bag,bai2019coordinate}. These features focus on the visual content of the images. These features can be further divided into two types: traditional content features ~\cite{zeglazi_sift_2016,oliva2005gist,oliva_modeling_2001,dalal2005histograms,lazebnik2006beyond,wu_centrist:_2011,xiao_mcentrist:_2014,margolin2014otc,quattoni_recognizing_2009,zhu_large_2010,li2010object,parizi2012reconfigurable,juneja2013blocks,lin_learning_2014,shenghuagao2010local,perronnin2010improving} and deep learning-based content features~\cite{gong_multi-scale_2014,kuzborskij2016naive,he2016deep,zhou2016places,zhang2017image,tang_g-ms2f:_2017,8085139,guo2016bag,bai2019coordinate}.

Most of the traditional content features are computed based on the popular traditional methods 
such as Scale-Invariant Feature Transform (SIFT)~\cite{zeglazi_sift_2016,peng2012finger},
Histogram of Gradient (HOG)~\cite{dalal2005histograms},
Generalized
Search Trees (GIST)~\cite{oliva2005gist,oliva_modeling_2001},
Spatial Pyramid Matching (SPM)~\cite{lazebnik2006beyond},
CENsus TRansform hISTogram (CENTRIST)~\cite{wu_centrist:_2011}, multi-channel (mCENTRIST)~\cite{xiao_mcentrist:_2014}, 
Oriented Texture Curves (OTC)~\cite{margolin2014otc},
GIST-Color ~\cite{oliva_modeling_2001},
RoI (Region of Interest) with GIST ~\cite{quattoni_recognizing_2009},
Max margin Scene (MM-Scene)~\cite{zhu_large_2010},
Object bank ~\cite{li2010object},
Reconfigurable BoW (RBoW) ~\cite{parizi2012reconfigurable},
Bag of Parts (BoP) ~\cite{juneja2013blocks},
Important Spatial Pooling Region (ISPR) ~\cite{lin_learning_2014},
Laplacian Sparse Coding SPM (LscSPM) ~\cite{shenghuagao2010local},
Improved Fisher Vector (IFV) ~\cite{perronnin2010improving}, and so on. 

In early works such as Search Trees GIST~\cite{oliva2005gist,oliva_modeling_2001}, CENTRIST~\cite{wu_centrist:_2011} and mCENTRIST~\cite{xiao_mcentrist:_2014}, features were extracted from local details such as colors, pixels, orientations, etc.
of the images. They have a limited ability to deal with the significant variations in the local content of images. Local features extracted by SIFT~\cite{zeglazi_sift_2016}, are multi-scale and rotation-invariant in nature. These local information may not provide global layout information of images and hence results in lower classification accuracy. SPM~\cite{lazebnik2006beyond} extracted the features based on the spatial regions of the image defined, by partitioning images into different slices. The features of each region were extracted as Bag of Visual Words (BoVW) of SIFT descriptors. Though their method captures some semantic regions of the image by partitioning method to some extent, it is not good at representing complex scene images that demand high-level information (object or scene details) for their better separability.  

In HoG~\cite{dalal2005histograms}, features were based on the gradient orientations in local grids of images. Since these features focus on designing histograms based on edge orientations, they are not suitable for complex scene images with multiple objects and their associations. In OTC~\cite{margolin2014otc}, features were based on the color variation of patches in images. These features are suitable to represent texture images but they may not be suitable for scene images. Similarly, several other methods such as \cite{oliva_modeling_2001,quattoni_recognizing_2009,zhu_large_2010,li2010object,parizi2012reconfigurable,juneja2013blocks,lin_learning_2014,shenghuagao2010local,perronnin2010improving} extracted features based on local sense of images. Since all these methods rely on the fundamental components and fail to capture the association/relations between components, they have limited capability for the classification of scene images. 

To sum up, these traditional content features are insufficient to represent the complex images such as scenes, where multiple factors such as context, background and foreground need to be considered for better differentiation.

Recent research works such as 
CNN-MOP~\cite{gong_multi-scale_2014},
CNN-sNBNL~\cite{kuzborskij2016naive},
ResNet152~\cite{he2016deep},
VGG~\cite{zhou2016places},
EISR~\cite{zhang2017image},
GM2SF~\cite{tang_g-ms2f:_2017},
Bag of Surrogate Parts (BoSP)~\cite{8085139,guo2016bag}, CNN-LSTM~\cite{bai2019coordinate}, and HDF \cite{sitaula2020hdf} used deep learning models to extract semantic features of images. Deep features are extracted from the intermediate layers of such pre-trained models and outperform existing traditional visual content features in scene image classification.

 Initially, Gong et al. 
 ~\cite{gong_multi-scale_2014} and Kuzborskij et al.~\cite{kuzborskij2016naive} extracted deep features from Caffe~\cite{jia2014caffe} model pre-trained on hybrid datasets (ImageNet~\cite{deng_imagenet:_2009} and Places~\cite{zhou2016places}) and ImageNet~\cite{deng_imagenet:_2009}, respectively. Specifically, Gong et al.~\cite{gong_multi-scale_2014} utilized fully connected layers ($FC$-layers) resulting in features size of $4,096$-D for each scale of the image to achieve orderless multi-scale pooling features. The size of final features is higher as the number of scales increases in their experiments. Their method outperforms single-scaled features though Gong et al. has a higher dimensional features size. 
 Similarly, Kuzborskij et al.~\cite{kuzborskij2016naive} also used $FC$-layers after fine-tuning. The features were used on top of Naive Bayes non-linear learning approach for the image classification purpose.  {Their method needs fine-tuning operations, which could need massive datasets for learning discriminating features.}
 
 Furthermore, He et al. \cite{he2016deep} proposed a deep architecture to extract features based on residual networks. The network was trained with ImageNet \cite{deng_imagenet:_2009} dataset, as with previous researchers in prior deep learning models such as VGG-Net \cite{simonyan2014very}. 
 Nevertheless, there was a necessity of deep architectures pre-trained with scene related images for the extraction of scene features.
Thus, Zhou et al.~\cite{zhou2016places} trained VGG-Net model \cite{simonyan2014very} with scene image dataset, which utilizes
the background information of the image. Their background features (scene-based) of scene images were more prominent (in terms of classification) than foreground features (object-based) obtained from ImageNet \cite{deng_imagenet:_2009} pre-trained models such as VGG16 \cite{simonyan2014very}.

Furthermore, some studies extracted the mid-level features based on the pre-trained deep learning models to improve the separability. Zhang et al.~\cite{zhang2017image} performed random cropping of images into multiple crops and extracted the visual features from the AlexNet~\cite{krizhevsky2012imagenet} model, pre-trained on ImageNet \cite{deng_imagenet:_2009}, to design a codebook of size of $1,000$. Then, sparse coding technique was used to extract the proposed features. The sparse coded features were concatenated with the tag-based features as the final features. Such types of features were extracted for each crop and concatenated as the final features of images. Their method suffers from the curse of the high features dimensionality. Furthermore, due to the chance of repetition of random patches, the chances of classification performance degradation is not surprising.

Tang et al.~\cite{tang_g-ms2f:_2017} chose three classification layers of GoogleNet \cite{szegedy2015going}, which was fine tuned with the scene images of corresponding domains. The features were extracted in the form of probabilities from classification layers and then performed feature fusion of these three probability-based features. Their method is also called multi-stage CNN (Convolutional Neural Network). {Their method demands large datasets and rigorous hyper-parameter tuning process to learn the separable features for the better discrimination of each input image.} 
Likewise, Guo et al.~\cite{8085139,guo2016bag} proposed Bag of Surrogate Parts (BoSP) features based on the higher pooling layers of the VGG16 model \cite{simonyan2014very} pre-trained with ImageNet \cite{deng_imagenet:_2009} such as $4^{th}$ and $5^{th}$ pooling layers.
Their method based on foreground information outperforms existing state-of-the-art methods.
However, their method simply focuses on the foreground (object-based) information because their deep learning model has been pre-trained on ImageNet \cite{deng_imagenet:_2009} with frequent object based information.

Bai et al. \cite{bai2019coordinate} extracted image features based on the Long Short Term Memory (LSTM) on top of Convolutional Neural Networks (CNNs). They assumed that the ordered slices of the images come under the LSTM problem in scene image representation, which yields prominent features. The features of each slice were extracted from the VGG16 \cite{simonyan2014very} pre-trained on Places \cite{zhou2016places} and fed into the LSTM model. The use of background information with such model outperformed several other previous methods including traditional methods and deep learning-based methods. {Since their method uses both CNN and LSTM models for the classification, it not only requires massive datasets to learn the highly separable features but also needs arduous hyper-parameter tuning process to mitigate both over-fitting and under-fitting problems. In addition, their method ignores the object-level information, which could be very important clues for better differentiation of complex scene images.} 

Recently, Sitaula et al. \cite{sitaula2020hdf} adopted the whole and part-level approach to represent the scene images. For this, they utilized both foreground and background information. From their experiment, they reported that the feature extraction using both foreground and background information at both whole level and part level can help capture the more interesting regions, which improves the class separability. { Since their method only relies on content information, it still may be unable to differentiate complex scene images having higher inter-class similarity problems.} 

Aggregation of multi-scale features provides more discriminating information of scene images by their multi-scale nature. However, this has been ignored by previous methods whose representation are often based on either foreground features or background features using pre-trained deep learning models. The fusion of such multi-scale foreground features and background features can help to provide the stable and improved classification accuracy of scene images even on varying sized images.

\begin{figure*}[t]
    \centering
    \includegraphics[width=0.95\textwidth, height=15cm,keepaspectratio]{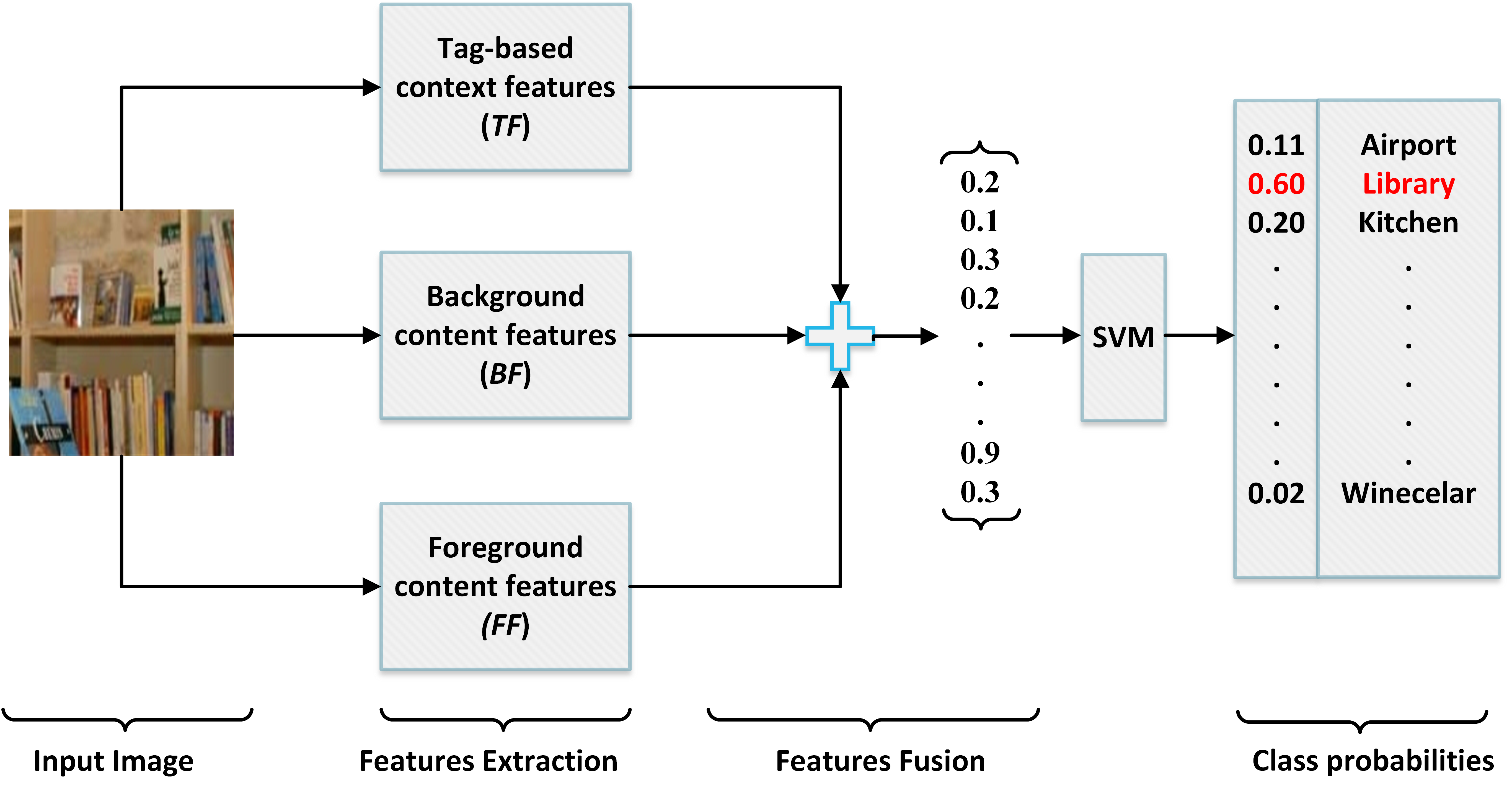}
    \caption{
    Overview of the proposed method.
    }
    \label{fig:2}
\end{figure*}

\subsection{Context features}

There are basically two recent works~\cite{zhang2017image,wang2019task} which used context information based on the users' annotation/description of images on the web. They fused such features with other visual features as the final representation of scene images for the classification purpose.

The early work in scene image representation that exploits the contextual features is Zhang et al.~\cite{zhang2017image}. It used the annotation/description tags of top $50$ visually searched images of the corresponding image and fused with visual features to represent the scene images. The description/annotation tags of all visually similar images
were retrieved by the internet tools (search engine) for each corresponding image query. However, their research involves a major problem. Their features size is extremely high because they use Bag-of-Words (BoW) of the raw tags extracted after pre-processing operations. The features size could be reduced heavily if they have used filter banks, which filter out the outlier tags present in the representing meta tags of the image. The exploitation of the filter banks not only reduces the dimension of the features but also improves the classification accuracy owing to the presence of its task-specific vocabularies.

Similarly, the shortage of Zhang et al.~\cite{zhang2017image} was fulfilled by Wang et al.~\cite{wang2019task}, who introduced the task-generic filter banks using pre-defined category names to filter out the outlier tags to some extent. To design the filter banks, they used pre-defined tags (category label) from two publicly available datasets: ImageNet~\cite{deng_imagenet:_2009} and Places~\cite{zhou2016places}. Based on the category labels of these datasets, outlier tags were filtered out to some degree.
Nevertheless, their method lacks domain-specific filters (e.g., scene images related tags) to work on the focused domain of research, and degrades the classification accuracy owing to the presence of less specific tags while designing filter banks.
Furthermore, their method suffers from out-of-vocabulary (OOV) tags while calculating semantic distance using WordNet~\cite{miller1995wordnet}. WordNet \cite{miller1995wordnet} cannot calculate the semantic distance of tags that are not present in the library. This results in the accumulation of some unnecessary tags in the filer banks belonging to WordNet \cite{miller1995wordnet} vocabularies only. This could ultimately degrade the classification accuracy using such tag-based context features.

The contextual information helps to leverage the users' knowledge about the scene images on the web. Such information can be important clues for discriminating ambiguous scene images with intra-class variation and inter-class similarity \cite{wang2019task,sitaula2019tag}
. In addition, content features are also found to have a better representation ability for the non-ambiguous images \cite{guo2016bag,8085139}. 
As such, it is necessary to integrate context and content information, to assist in dealing both types (ambiguous and non-ambiguous) information present in scene images. This could boost classification accuracy of images, especially for scene images.

\section{Our Approach}
\label{prop:method}

 In this section, we explain how to extract context and content-based features and integrate them for the task of scene image classification using Support Vector Machine (SVM) classifier. The overview of our proposed method is shown in Fig. \ref{fig:2}. Contextual information of scene images can be obtained from their annotations/descriptions. Annotations/descriptions for some images may not be available. Instead, we can exploit annotations/descriptions of similar images available in the internet. We propose to extract tag-based context features ($TF$) of an image from the annotations/descriptions of similar images on the internet. For content features, we propose to extract features from background ($BF$) and foreground ($FF$) information in images using deep learning models. Finally, we perform the fusion of all these context and content-based features ($CCF$). 

\subsection{Extraction of tag-based context features}

\begin{figure*}[t]
    \centering
    \includegraphics[width=0.95\textwidth, height=15cm,keepaspectratio]{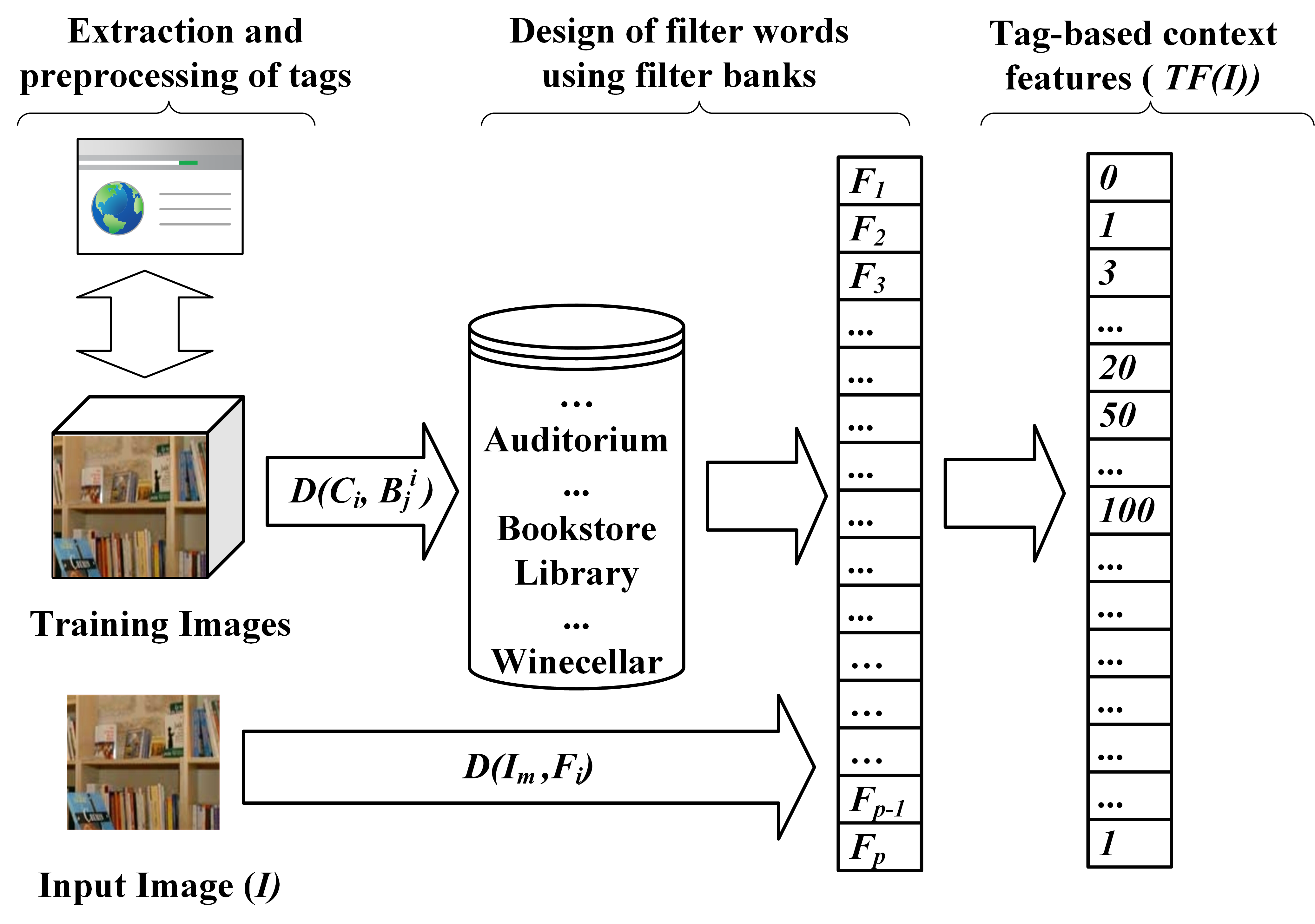}
    \caption{
    Block diagram to extract the tag-based context features. Here, $D(C_i, B_j^i)$ provides the averaged similarity between the category label ($C_i$) and the unique tags ($B_j^i$) 
    for the corresponding image
    . Similarly, $D(I_m, F_{i})$ yields the averaged similarity between $m^{th}$ tag of input image $I$ ($I_m$) and filter word ($F_{i}$) to extract our proposed tag-based context features ($TF(I)$ is a vector representation of image $I$) of the input image. In the block diagram, input image ($I$) represents the collection of pre-processed tags $I_m$. 
    }
    \label{fig:3}
\end{figure*}

We take four steps to extract tag-based context features of a scene image, as illustrated in Fig. \ref{fig:3}.

\subsubsection{Extraction of the annotation/descriptions of similar images}
For each scene image, we first extract its similar images using the Yandex\footnote{www.yandex.com} search engine as suggested by \cite{sitaula2019indoor}. We use top $50$ visually similar images for each input image, as suggested by previous works~\cite{wang2019task,wang2019learning,sitaula2019tag}. 
Then, we collect the descriptions/annotations of all $50$ images to yield the raw representative tags for the queried input image.

\subsubsection{Pre-processing of annotation/description tags}\label{subsubsec_tagsPreprocessing}
The extracted annotation/descriptions are pre-processed using text mining approach, which includes language translation (for some texts that were in Russian Language), tokenization, stop words removal, and redundant token removal. We capture the unique tokens among the redundant tokens such as plane, planes, snowboarding, snowboard, etc. otherwise, they may negatively impact the classification performance. Over-stemmed and under-stemmed tags do not function properly in our work because word embeddings that we use prefers meaningful tags to achieve better embedding vectors. We observe that traditional Stemmer and Lematizer produce meaningless tags in our work by over-stemming and under-stemming most of the time, which became a major obstacle in our work.

To mitigate this problem, we design an algorithm that outputs unique meaningful tags based on the traditional Lematizers and Stemmers. Here, Alg. \ref{algo:0} outputs unique tags ($B$) from raw tags ($R$) through exploiting the idea of string sorting technique \cite{bentley1997fast}. The raw tags ($R$) are initially lematized and stemmed ($S$) to achieve corresponding root of the raw tags for grouping similar raw tags in the algorithm. Then, we sort the strings to find the possible unique tags. Note that we only utilize Alg. \ref{algo:0} to achieve unique tags in the filter words (codebook), which results in reduced size of codebook in our method.

\begin{algorithm}[t]
 \caption{Unique tags extraction from raw tags}
 \begin{algorithmic}[1]
 \renewcommand{\algorithmicrequire}{\textbf{Input:}}
 \renewcommand{\algorithmicensure}{\textbf{Output:}}
 \REQUIRE $R \leftarrow$ Raw tags,\\ 
 $S \leftarrow$ Corresponding lematized and stemmed raw tags
 \ENSURE  $B$ \algorithmiccomment{Unique tags},
 \STATE $B \leftarrow {[]}$ \algorithmiccomment{Empty unique tags holder}
  \STATE $N = |R|$ \algorithmiccomment{Size of tag lists}
  \FOR {$i = 0$ to $N$}
   \STATE $T \leftarrow {[]}$ \algorithmiccomment{Temporary list}
   \FOR {$j = 0$ to $N$}
     \IF{$S_i$==$S_j$}
       \STATE $T.add(R_j)$ \algorithmiccomment{Add the corresponding raw tag} 
     \ENDIF
  \ENDFOR
  \STATE $T \leftarrow SORT(T)$ \algorithmiccomment{Sort in the ascending order}
     \STATE $B.add(T[0])$ \algorithmiccomment{Add the first raw tag}
  \ENDFOR
  \STATE $B \leftarrow UNIQUE(B)$ \algorithmiccomment{Remove duplicate tags}
 \RETURN $B$
 \end{algorithmic}
 \label{algo:0} 
 \end{algorithm}
 
\begin{table}
\caption{Example filter banks of some categories retrieved by our method on the MIT-67 dataset. }
\centering
\vspace{4mm}
\begin{tabular}{p{2.5cm}|p{7cm}}
\toprule
Category & Filter banks\\
\midrule
Airport inside & airport, inside, airline, flight, hotel, plane, aircraft, terminal, etc.\\
\hline
Library & libraries, librarian, bookstore, musician, bookshop, collections, classroom, archive, etc.\\
\hline
Winecellar & cellar, wines, tasting, grapes, beer, wineries, whiskey, vino, basement, winemaker, vineyard, brandy, etc.\\
\hline
Subway & subway, tram, train, transit, metro, commuter, bus, escalator, etc.\\
\bottomrule
\end{tabular}
\label{tab:0}
\end{table} 

\begin{figure*}[t]
    \centering
    \includegraphics[width=0.95\textwidth, height=15cm,keepaspectratio]{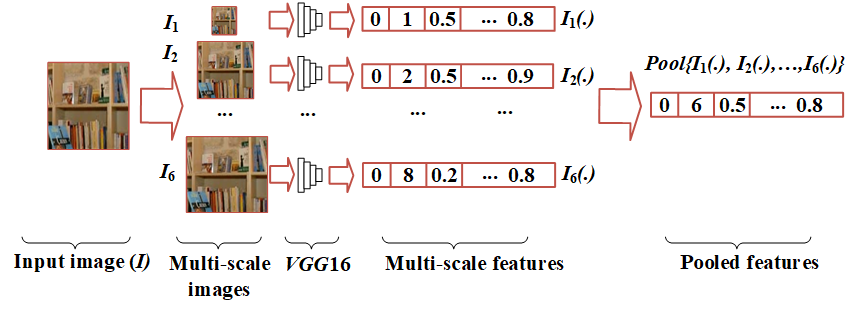}
    \caption{Block diagram to extract the content features (foreground and background) of the input image, $I$. Here, $I_1(.)$, $I_2(.),\cdots,I_6(.)$ represent the global average pooling (GAP) features extracted from $VGG16$ for $I_1$ to $I_6$ images at six different scales. We use two types of $VGG16$, $VGG16_{Places}$ and $VGG16_{ImageNet}$ which yield $I_i(s)$ and $I_i(o)$, respectively. }
    \label{fig:4}
\end{figure*}

\subsubsection{Design of filter words (codebook)}
With the training images, we design filter banks and extract filter words. To achieve this, we first select top $500$ frequent words from the raw tags of training images from each category $i$ ($i=1,2,\cdots,n$, where $n$ is the number of categories in a given dataset) \cite{sitaula2019tag}. Then, we calculate the average similarity between the category label $C_i$ and these top frequent words. For this, we need to represent words as vectors. For vector representations of words, we utilize three different types of word embeddings vectors, namely, Word2Vec \cite{mikolov2013efficient}, GloVe \cite{pennington2014glove}, and fastText \cite{bojanowski2017enriching}, which leverage the semantic knowledge from three different domains.
Since we utilize three types of word embeddings to show the relationship between a category label and its highly frequent words, we average three cosine similarity scores as the final semantic similarity of a tag to the corresponding category label. We repeat such procedure for all $500$ highly frequent tags for category $i$. Finally, we select top-$k$ most similar tags to form the filter bank ($F_i$) for category $i$ (see an example in Table \ref{tab:0}). We repeat the same process for all $n$ categories. We then take the union of all $n$ filter banks removing duplicates to achieve the final filter words (codebook) $F=\cup_i F_i$, which is used to extract tag-based context features of images. The size of the context features is $|F|\leq n\times k$ (let's say $|F|=p$). Empirically, we found that $k=25$ is a good setting (see  Section~\ref{topk}). Therefore, we use $k=25$ on all datasets.

Mathematically, $D(C_i, B_j^i)$ be the average of cosine similarities between a category label $C_i$ and a frequent word $B_j^i$ using Word2Vec ($wv$), GloVe ($gv$), and fastText ($ft$) word embedding techniques (Eq. \eqref{eq:1}). Let $r(\cdot)$ be the vector representation of a word using word embedding technique $r\in\{wv,gv,tf\}$.
\begin{equation}
\label{eq:1}
 D(C_i, B_j^i) = \frac{1}{3} \sum_{r\in\{wv,gv,ft\}} cos\left(r(C_i), r(B_j^i)\right)\textcolor{red}{,}
\end{equation}

The cosine similarity of two vectors $k_1$ and $k_2$ is defined as follows:
\begin{equation}
\label{eq:2}
 cos(k_1, k_2) = \frac {k_1 \cdot k_2}{||k_1|| \cdot ||k_2||} \textcolor{red}{,} 
\end{equation}

\subsubsection{Extraction of tag-based context features}
This is the final step to extract the proposed tag-based context features ($TF$) shown in Fig. \ref{fig:3}. The tag-based context features are based on a histogram of tags present in the filter words $F$. Let $TF(I)$ be the vector representation of image $I$ using tag-based features. Note that $|TF(I)|$ is the same as $|F|$. $TF(I)$ is computed as follows.
Initially, the pre-processed tags from annotations/descriptions of $50$ similar images of $I$ on the web are achieved.
We believe that normal string matching of pre-processed tags of the image and those unique filter word in the codebook is insufficient to capture semantic relationship between tags in generating histogram. Thus, we use the average cosine similarity (Eq.~\ref{eq:1}) threshold to match image tags semantically related to codebook tags. For this, we use a fixed threshold of $\Lambda= 0.40$ suggested by Sitaula et al. \cite{sitaula2019tag}. 
The procedure to generate $TF(I)$ of image $I$ from its pre-processed tags is provided in Alg.~\ref{algo:2}. 

\begin{algorithm}[t]
 \caption{Extraction of tag-based features of an image}
 \begin{algorithmic}[1]
 \renewcommand{\algorithmicrequire}{\textbf{Input:}}
 \renewcommand{\algorithmicensure}{\textbf{Output:}}
\REQUIRE $I \leftarrow$ Pre-processed tags for an image $I$,
 $F \leftarrow$ Unique filter words (codebook)
 \ENSURE $TF(I)$ \algorithmiccomment{$p$-sized tag-based features of image $I$}
  \STATE Set all $p$ elements of $TF(I)$ to 0 \algorithmiccomment{Initialisation}
  \FOR {$m = 0$ to $|I|$}
   \FOR {$i = 0$ to $|F|$}
     \IF{$D(I_m,F_i)\geq \Lambda$}
       \STATE $TF_i(I)++$ \algorithmiccomment{Increment the respective bin count} 
     \ENDIF
  \ENDFOR
  \ENDFOR
 \RETURN $TF(I)$
 \end{algorithmic}
 \label{algo:2} 
 \end{algorithm}

\subsection{Extraction of content features}

For content-based features, we suggest to extract deep features based on background and foreground information in images using pre-trained deep learning models which can provide useful content information. We use the VGG16 model \cite{simonyan2014very} pre-trained with ImageNet~\cite{deng_imagenet:_2009} and Places~\cite{zhou2017places} to extract the foreground and background content features, respectively. We prefer VGG16 for {three} reasons. Firstly, it has a simple architecture, yet prominent feature extraction capability, especially for the scene image representation, as suggested by recent previous works {related to scene images} \cite{guo2016bag,8085139,bai2019coordinate,sitaula2020hdf}. Secondly, it has only five pooling layers that makes us easier to analyze and evaluate the image representation than other complex models having several layers such as GoogleNet \cite{szegedy2015going}, Inception-V3 \cite{szegedy_rethinking_2016}, etc. {Finally, VGG16 has shown the best top-1 accuracy compared to AlexNet and GoogleNet on scene image representation using background information \cite{zhou2017places}, where only three deep learning models (VGG16, GoogleNet, and AlexNet) have been pre-trained with Places datasets for the extraction of background information.} 

Before inputting each image into the pre-trained models, we resize them to $512\times512$ pixels and perform feature extraction. A higher pooling layers such as the $5^{th}$ of VGG16 model \cite{simonyan2014very} is found to have highly separable features than other intermediate layers \cite{8085139,guo2016bag} in scene image representation. Thus, we utilize the $5^{th}$ pooling layer to extract the content features and perform global average pooling (GAP) operation on them to achieve 512-D features size. To capture the content information of images at different scales, we vary the resolution of each resized image at 6 different scales $\tau \in \{0.6, 0.8, 1.0, 1.2, 1.4, 1.6\}$. The features achieved from multiple scale levels are aggregated to achieve multi-scale features. We perform the aggregation process on both foreground content features ($FF$) and background content features ($BF$) separately. Eqs. \eqref{eq:5} and \eqref{eq:6} extract the background and foreground features of the input image $I$, respectively.
We use VGG16 pre-trained on Places ($VGG16_{Places}$) and VGG16 pre-trained on ImageNet ($VGG16_{ImageNet}$) to extract the background and foreground features, respectively.
Furthermore, $I_a$ (where $a=1,2,\cdots,6$) represents the image at the $a^{th}$ scale of the input image ($I$) and the background ($I_a(s)$) and foreground features ($I_a(o)$) are extracted at the corresponding scale.
The aggregation of all $I_a(s)$ features from multi-scale images achieves multi-scale background features ($BF(I)$), and also the aggregation of all $I_a(o)$ features from multi-scale images provides multi-scale foreground features ($FF(I)$). To achieve them, we use the $Pool(.)$ function, which aggregates multi-scale features of the input image ($I$) on its both types of content features (foreground and background). Empirically, we found that max pooling produces the best result (see Section~\ref{multiscale}). Therefore, we use max pooling aggregation to achieve multi-scale features.
\begin{equation}
\label{eq:5}
{BF(I)} = Pool\{I_1(s), I_2(s),\cdots, I_6(s)\},
\end{equation}
where $I_a(s)=VGG16_{Places}(I_a)$
\begin{equation}
\label{eq:6}
{FF(I)} =Pool\{I_1(o), I_2(o),\cdots, I_6(o)\},
\end{equation}
where $I_a(o)=VGG16_{ImageNet}(I_a)$.

The block diagram for the extraction of such features is shown in Fig. \ref{fig:4}. We adopt this pipeline to extract the background ($BF(I)$) and foreground ($FF(I)$) features of the input image ($I$) separately.

\subsection{Fusion of content and context features}

To achieve the final features, we use a simple yet prominent serial (concatenation) feature fusion method \cite{yang2003feature} of all these three different types of features (Eq. \eqref{eq:7}): $FF$ and $BF$ represent the content features and $TF$ represents the context features) {although there are several feature fusion algorithms in literature \cite{gad2019iot,peng2015linear,jing2014saliency}. } 
It should be noted that the features size of the context-based features ($TF$) is not fixed, and it can be higher/lower than the content-based features ($BF$ and $FF$). Concatenating features of unequal sizes could create bias to certain types of features for the classification. Thus, we reduce the features size of higher dimensional features to the minimum size of all types of features using Principal Component Analysis (PCA). In particular, if the size of the $TF(I)$ features is higher than others such as $FF(I)$ and $BF(I)$, we reduce its size into the minimum size among the three types of features. This ensures all types of features have the same size, which avoids the bias to higher sizes of features. Finally, we concatenate all three same-sized features to achieve the final proposed features for the classification task. We assume that the concatenation method here could help to preserve the information achieved from different domains because min, max, or avg methods may not be sufficient to capture all the corresponding information in such a case.

\begin{equation}
\label{eq:7}
{CCF(I)} =\{TF(I), BF(I), FF(I)\},
\end{equation}

{
\subsection{Computational complexity analysis}
In this subsection, we study the computational complexity of both  content features and context (tag-based) features. First, we analyze the computational complexity of the proposed content features based on two VGG16 models using FLOPs (FLoating-point OPerations). 
FLOPs is the multiplication of spatial width of the map, spatial height of the map, previous layer depth, current layer depth, kernel width, kernel height, and repeats. Since we have used six different scales \{0.6, 0.8, 1.0, 1.2, 1.4, 1.6\} based on fixed image size ($512 \times 512$ pixels), we add FLOPs required from all six different image sizes in pixels ($307 \times 307$, $409 \times 409$, $512 \times 512$, $614 \times 614$, $716 \times 716$, $819 \times 819$) for the total FLOPs. Specifically, the image of scales 0.6, 0.8, 1.0, 1.2, 1.4, and 1.6 impart $0.5E+09$, $1.0E+09$, $1.8E+09$, $2.5E+09$, $3.4E+09$, and $4.4E+09$ FLOPs, respectively, resulting in the total of $13.6E+09$ FLOPs for all scales.  Moreover, we have used two VGG16 models (for both foreground and background information extraction), which yield $2 \times (13.6E+09)$ FLOPs. Second, we analyze the computational complexity of the proposed context features using time complexity. For this, we consider two main algorithms (Algs. \ref{algo:0} and \ref{algo:2}). Here, Alg. \ref{algo:0}, which is used for the extraction of unique tags, imparts $O(N^2)$ complexity including sorting operation, where $N$ denotes the total number of tokens in the document representing the input image in our work. In addition, Alg. \ref{algo:2}, which is used for the extraction of context features, provides $O(I\times F)$ complexity, where $I$ and $F$ represent the length of pre-processed document representing the image and length of the supervised codebook, respectively.  
}

\section{Experimental setup and Analysis}
\label{experiments}
\subsection{Dataset}

We evaluate the performance of the introduced three types of features -- context-based features ($TF$), deep content-based features ($FF+BF$) and the fusion of context- and content-based features, in the scene image classification task using Support Vector Machine (SVM) classifier. We compare our features with a wide range of traditional computer vision (CV)-based content features, deep learning-based content features and tag-based context features. We use three commonly-used scene image datasets: MIT-67~\cite{quattoni_recognizing_2009}, Scene-15~\cite{fei-fei_bayesian_2005}, and Event-8~\cite{li2007and}.

\begin{figure}[t]
\begin{center}
 \includegraphics[width=0.90\textwidth, height=50mm,keepaspectratio]{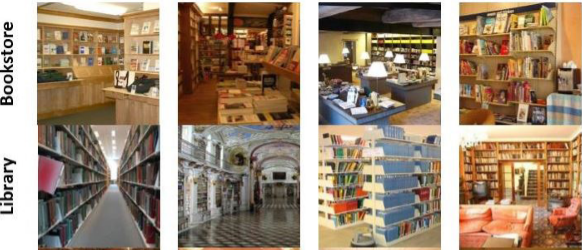}
  \caption{Example images from the MIT-67 dataset \cite{quattoni_recognizing_2009}. 
  }
  \label{fig:mit}
 \end{center}
  \end{figure}

{\bf MIT-67} is the largest indoor scene dataset containing $15,620$ images belonging to $67$ categories, with at least $100$ images per category. Most of the images on MIT-67 contains indoor contents with varying complexities such as inter-class similarities and intra-class dissimilarities. Example images of MIT-67 are shown in Fig. \ref{fig:mit}. For the experiments, we follow the protocol ~\cite{quattoni_recognizing_2009} with $80/20$ train/test split.
Some researchers \cite{quattoni_recognizing_2009,zhu_large_2010, li2010object,parizi2012reconfigurable,juneja2013blocks, margolin2014otc,lin_learning_2014,zhang2017image,gong_multi-scale_2014, zhou2016places,he2016deep,8085139,guo2016bag,tang_g-ms2f:_2017, bai2019coordinate,kim2014convolutional,wang2019task} used such setup in their experiments for the scene images representation and classification purpose.

\begin{figure}[t]
\begin{center}
 \includegraphics[width=0.90\textwidth, height=50mm,keepaspectratio]{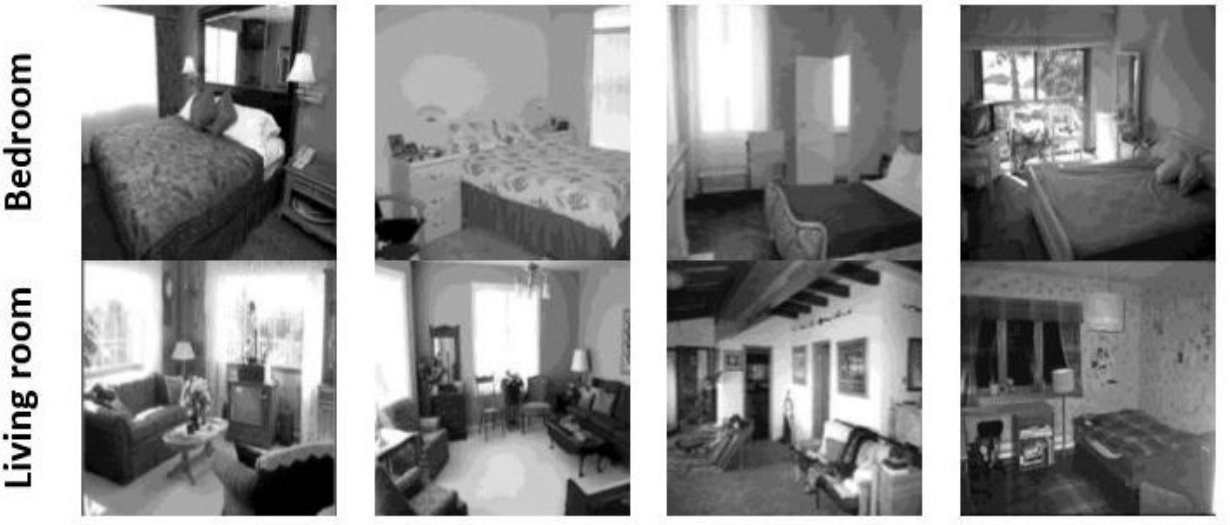}
  \caption{Example images from the Scene-15 dataset \cite{fei-fei_bayesian_2005}. }
  \label{fig:scene15}
 \end{center}
  \end{figure}

{\bf Scene-15} contains $4,485$ images corresponding to $15$ categories, each category comprising $200$ to $400$ images. Some example images are shown in Fig. \ref{fig:scene15}.
To perform our experiments, we design $10$ sets of train/test splits and reported the average accuracy. For this, we select $100$ images/category for training and remaining images as testing in each set, as done by previous studies~\cite{oliva_modeling_2001,lazebnik2006beyond,wu_centrist:_2011,margolin2014otc,lin_learning_2014,perronnin2010improving,zhou2016places,zhang2017image,tang_g-ms2f:_2017,he2016deep,kim2014convolutional,wang2019task}.

\begin{figure}[t]
\begin{center}
 \includegraphics[width=0.90\textwidth, height=50mm,keepaspectratio]{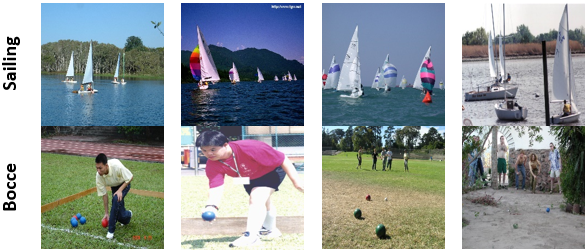}
  \caption{Example images from the Event-8 dataset \cite{li2007and}. }
  \label{fig:event8}
 \end{center}
  \end{figure}

{\bf Event-8} includes $8$ different categories and $1,579$ images in total, where each category contains $137$ to $250$ images. It is a collection of outdoor images of various sporting events. Some example images from the dataset are provided in Fig. \ref{fig:event8}. we design $10$ sets of train/test splits and report the average accuracy. For each set, we separate $70$ images per category as a training set and $60$ images as a testing set, as employed in previous studies ~\cite{li2010object,lin_learning_2014,shenghuagao2010local, perronnin2010improving,zhang2017image,kuzborskij2016naive,zhou2016places, he2016deep,wang2019task,kim2014convolutional}.

\subsection{Implementation}
\label{implementation}

To implement our proposed framework, we use Pymagnitude~\cite{patel2018magnitude}, Keras ~\cite{chollet2015keras}, and Sklearn \footnote{https://scikit-learn.org/stable/} in Python. Pymagnitude is used to extract the word embeddings for the extraction of proposed tag-based context features. Similarly, Keras packages is used to implement the pre-trained deep learning models \cite{simonyan2014very, gkallia2017keras_places365} for the extraction of our content features. The content features are encoded and normalized as suggested by \cite{guo2016bag,8085139}. To avoid the divide-by-zero exception, we add $\epsilon=1e-7$ 
to the denominator during normalization. We standardize the content and context features before
the fusion to achieve the final proposed features. Finally, Sklearn is used to implement Support Vector Machine (SVM) for classification. To train a SVM, we fix $kernel=rbf$ and tune the $C$ values in the range of $\{1, 2,\cdots, 200, 1\times10^3, 2\times10^3,\cdots,6\times10^3\}$ 
and $gamma$ values in the range of $\{10^{-1}, 10^{-2},\cdots, 10^{-7}\}$.

\begin{table}
\caption{Classification accuracy of our proposed features and other existing features on the MIT-67 dataset. 
The best result is in bold.}
\vspace{1mm}
\centering
\begin{tabular}{p{7cm} p{2cm}}
\toprule
Method & Accuracy(\%)\\
\midrule
Traditional Computer vision-based methods\\
\midrule
ROI with GIST, 2009 \cite{quattoni_recognizing_2009} &26.1 \\
MM-Scene, 2010 \cite{zhu_large_2010} & 28.3\\
Object Bank, 2010 \cite{li2010object}&37.6\\
RBoW, 2012 \cite{parizi2012reconfigurable}&37.9\\
BOP, 2013 \cite{juneja2013blocks}&46.1\\
OTC, 2014 \cite{margolin2014otc}&47.3\\
ISPR, 2014 \cite{lin_learning_2014}&50.1\\
\midrule
Deep learning-based methods\\
\midrule
EISR, 2017 \cite{zhang2017image} &66.2\\
CNN-MOP, 2014 \cite{gong_multi-scale_2014} & 68.0\\
VGG, 2016 \cite{zhou2016places} & 75.3 \\
ResNet152, 2016 \cite{he2016deep} & 77.4\\
SBoSP-fusion, 2016 \cite{guo2016bag} &77.9\\
G-MS2F, 2017 \cite{tang_g-ms2f:_2017}& 79.6\\
BoSP-Pre\_gp, 2018 \cite{8085139} & 78.2 \\
CNN-LSTM, 2019 \cite{bai2019coordinate}& 80.5\\
{HDF, 2020} \cite{sitaula2020hdf}&82.0 \\
\midrule
Tag-based methods\\
\midrule
CNN, 2014 \cite{kim2014convolutional} & 52.0\\
BoW, 2019 \cite{wang2019task} &52.5\\
s-CNN(max), 2019 \cite{wang2019task} & 54.6\\
s-CNN(avg), 2019 \cite{wang2019task} & 55.1\\
s-CNNC(max), 2019 \cite{wang2019task} & 55.9\\
{TSF, 2019} \cite{sitaula2019tag} &76.5 \\
\hline
\textbf{Ours \emph{TF}}&77.1 \\
\textbf{Ours \emph{DF (FF+BF)}}&82.0 \\
\textbf{Ours \emph{CCF}} &\textbf{87.3} \\
\hline
\end{tabular}
\label{tab:mit67}
\end{table} 

\subsection{Comparison with the state-of-the-art methods}

We compare the SVM classification performance of our three types of features - tag-based context features ($TF$), deep content features ($DF=FF+BF$) and fusion of context and content features ($CCF$), with the state-of-the-art feature extraction methods including traditional CV-based methods, deep learning-based methods and tag-based methods. We present the accuracy numbers of all contending measures on MIT-67, Scene-15, and Event-8 in three separate Tables \ref{tab:mit67}, \ref{tab:scene15} and \ref{tab:event8}, respectively. For existing methods, we simply present the accuracies reported in corresponding published papers.

In Table \ref{tab:mit67}, we present the classification accuracies of all contending methods on the MIT-67 dataset. We compare our features with different types of previous methods: $7$ traditional CV-based methods, $9$ deep learning-based methods and $6$ tag-based methods. Our tag-based context features ($TF$) outperform all existing traditional CV-based methods and existing tag-based features, by at least $27\%$ and $0.6\%$, respectively. 
Our features produce competitive results to many state-of-the-art deep content-based features. Our deep features ($DF=FF+BF$) based on background and foreground information surpass all contending deep features by at least $1.5\%$ except HDF method. The fusion of the two types of features ($CCF$) further boost the performance by $5.3\%$ 
aga

t recent second best method (HDF \cite{sitaula2020hdf})
and leads to the classification accuracy of $87.3\%$. 

\begin{table}
\caption{Comparison of the proposed features' accuracy with the existing features on Scene15 dataset. The best result is in bold. }
\vspace{1mm}
\centering
\begin{tabular}{p{7cm} p{2cm}}
\toprule
Method & Accuracy(\%)\\
\midrule
Traditional Computer vision-based methods\\
\midrule
GIST-color, 2001 \cite{oliva_modeling_2001} &69.5\\
SPM, 2006 \cite{lazebnik2006beyond} &81.4 \\
IFV, 2010 \cite{perronnin2010improving}&89.2\\
CENTRIST, 2011 \cite{wu_centrist:_2011}&83.9\\
OTC, 2014 \cite{margolin2014otc}&84.4\\
ISPR, 2014 \cite{lin_learning_2014}&85.1\\
\midrule
Deep learning-based methods\\
\midrule
VGG, 2016 \cite{zhou2016places} & 93.0 \\
ResNet152, 2016 \cite{he2016deep}& 92.4\\
G-MS2F, 2017 \cite{tang_g-ms2f:_2017}&92.9\\
EISR, 2017 \cite{zhang2017image} &94.5\\
{HDF, 2020} \cite{sitaula2020hdf}& 93.9\\
\midrule
Tag-based methods\\
\midrule
CNN, 2014 \cite{kim2014convolutional} & 72.2\\
BoW, 2019 \cite{wang2019task} &70.1\\
s-CNN(max), 2019 \cite{wang2019task} &76.2 \\
s-CNN(avg), 2019 \cite{wang2019task} & 76.7\\
s-CNNC(max), 2019 \cite{wang2019task} & 77.2\\
{TSF, 2019} \cite{sitaula2019tag}& 81.3 \\
\hline
\textbf{Ours \emph{TF}}&84.9 \\
\textbf{Ours \emph{DF (FF+BF)}}&93.5\\
\textbf{Ours \emph{CCF}} & \textbf{95.4} \\
\hline
\end{tabular}
\label{tab:scene15}
\end{table}

Table \ref{tab:scene15} summarizes the classification accuracies on the Scene-15 dataset. For comparison with existing methods, we choose {$6$} traditional CV-based methods, {$5$} deep learning-based methods and $6$ tag-based methods. Our context ($TF$) features outperform all existing tag-based context features and some traditional CV-based methods, but they perform worse than other traditional CV-based and deep learning-based content features. Our deep features ($DF$) surpass all existing tag-based, traditional CV-based and deep features except EISR \cite{zhang2017image} which is slightly higher than our $DF$ by $1\%$. As a result, the fusion of our context and content-based features ($CCF$) exceed the best existing method (EISR \cite{zhang2017image}) by about $1\%$.

\begin{table}
\caption{Comparison of the proposed features' accuracy with the existing features on Event8 datast. The best result is in bold. }
\vspace{4mm}
\centering
\begin{tabular}{p{7cm} p{2cm}}
\toprule
Method & Accuracy(\%)\\
\midrule
Traditional Computer vision-based methods\\
\midrule
Object Bank, 2010 \cite{li2010object}& 76.3\\
LscSPM, 2010 \cite{shenghuagao2010local}&85.3\\
IFV, 2010 \cite{perronnin2010improving}&90.3\\
ISPR, 2014 \cite{lin_learning_2014}&74.9\\
\midrule
Deep learning-based methods\\
\midrule
CNN-sNBNL, 2016 \cite{kuzborskij2016naive} &95.3 \\
VGG, 2016 \cite{zhou2016places} & 95.6 \\
ResNet152, 2016 \cite{he2016deep} & 96.9 \\
EISR, 2017 \cite{zhang2017image} &92.7\\
{HDF, 2020} \cite{sitaula2020hdf}&96.2 \\
\midrule
Tag-based methods\\
\midrule
CNN, 2014 \cite{kim2014convolutional} &85.9 \\
BoW, 2019 \cite{wang2019task} &83.5\\
s-CNN(max), 2019 \cite{wang2019task} & 90.9\\
s-CNN(avg), 2019 \cite{wang2019task} & 91.2\\
s-CNNC(max), 2019 \cite{wang2019task} & 91.5\\
{TSF, 2019} \cite{sitaula2019tag} & 94.4\\
\hline
\textbf{Ours \emph{TF}}& 95.8 \\
\textbf{Ours \emph{DF (FF+BF)}}& 97.5\\
\textbf{Ours \emph{CCF}} &\textbf{98.1} \\
\hline
\end{tabular}
\label{tab:event8}
\end{table} 

In Table \ref{tab:event8}, we list the accuracies of contending feature extraction methods on Event-8. We select $4$ traditional CV-based methods, $5$ deep learning-based methods and $6$ tag-based methods.
The results show that our context feature ($TF$) outperforms all existing methods except ResNet152\cite{he2016deep}, whereas our deep content features ($DF$) outperform them all. The fusion of $TF$ and $DF$ ($CCF$) generates the best result, which surpasses the best existing method, ResNet152 \cite{he2016deep} by more than $1\%$. 

To summarise, our context features ($TF$) outperform existing context features on all three datasets, and our deep content features ($DF$) produce generally better (with an exception of EISR \cite{zhang2017image} on Scene15). However, by simply fusing the two types (context and content) of features ($CCF$), we achieve the {\bf best} results in all datasets with the margin of $0.9\%$ to $5.3\%$ over the best existing methods. It shows that fusing context information with the content information can better differentiate images of different classes. It adds more value (more than $5\%$ improvement in the classification accuracy) on the MIT-67 dataset, the largest dataset in our research involving lots of ambiguities between classes with inter-class similarity and intra-class dissimilarity. 

It is interesting to note that all three best existing methods -- CNN-LSTM \cite{bai2019coordinate} (MIT-67), EISR \cite{zhang2017image} (Scene-15) and ResNet-152 \cite{he2016deep} (Event-8), have greater features sizes, which is one major problem in many existing methods as well. For instance, CNN-LSTM \cite{bai2019coordinate}
has the feature size of 4,096-D on the MIT-67 dataset. Similarly, EISR \cite{zhang2017image} has an extremely higher dimensional features size, which is greater than $50\times 2,048$-D after concatenation with the the tag-based features. Finally, ResNet-152 \cite{he2016deep} yields a feature size of $2,048$-D, which is the average pooling layer of ResNet-152 and it is just the second best existing method on the Event-8 dataset.
In terms of the size of our features, we have the least features size than all other features. The size of our fused features ($CCF$) on the MIT-67, Scene-15 and Event-8 are $1,536$-D, $<811$-D and $<475$-D, respectively. The features size of images on the Scene-15 and Event-8 is not fixed because we used $10$ sets of train/test splits for each dataset and each set in the corresponding dataset has varied length of filter words, thus leading to varied sizes of tag-based context features. This impacts the final features size.
Also, the features size reduction through PCA to match the smallest size assists to decrease the final features size. Thus, our features size is significantly lower than existing methods' features size. Despite this, our features can still enable better classification performance than existing methods, since our fused features account for both context and content information. 

\begin{table}
\caption{Averaged classification accuracy (\%) with different  $k$ values for $top_k$ similar filter banks extraction for the tag-based context features ($TF$) on Event-8 dataset. The best result is in bold. }
\vspace{0mm}
\label{tab:topk}
\centering
\begin{tabular}{c| c| c| c| c| c|c}
\toprule
$k$ &15 & 25 & 50 & 75 & 100 & 125\\
\midrule
{Accuracy}&95.7& \textbf{95.8} & 95.6&95.5 &95.2 &94.8 \\
\bottomrule
\end{tabular}
\end{table} 

\subsection{Analysis of top $k$ in tag-based context features}
\label{topk}

To study the effect of the $k$ parameter on $top_k$ filter banks while extracting tag-based context features, we tune different values on Event-8 dataset and find the best $k$. We experiment different values of $k\in\{15,25,50,75,100,125\}$ in all $10$ sets on Event-8 and report the averaged accuracy in Table \ref{tab:topk}. Finally, we employ the best $k=25$ on remaining datasets for the tag-based context features extraction.

\subsection{Analysis of three individual features and their fusion}
We perform experiments on the Event-8 dataset to analyze the effect of the content and context features in the classification accuracy. We analyze the performance of the two types of content features (foreground features ($FF$) and background features ($BF$)) separately, and their aggregation
($FF+BF$, the proposed content features). Also, we analyze the tag-based context features ($TF$) only and the fusion of our context and content features ($CCF$).
The detailed results are listed in Table \ref{tab:content_context}.
We observe that our aggregated visual features, $FF+BF$ provide the second highest accuracy ($97.5\%$) following after $CCF$, which generates a classification accuracy of \textbf{$98.1\%$}. This infers that the two different types of information are found to have better separability in the scene images representation.

\begin{table}
\caption{Averaged classification accuracy (\%) for different combinations of proposed features on the Event8 dataset. The best result is in bold.}
\label{tab:content_context}
\vspace{4mm}
\centering
\begin{tabular}{p{1.5cm}| p{0.7cm}| p{0.7cm}|p{1cm}|p{1cm}| p{0.7cm}| p{2cm}|p{0.6cm}}
\toprule
 Features &$FF$ & $BF$ & $TF$& $BF+TF$ &
 $FF+TF$ &$DF$($FF+BF$) & $CCF$\\
\midrule
Accuracy &97.2&95.8 &95.8&97.0 &97.8 &97.5 &\textbf{98.1} \\
\bottomrule
\end{tabular}
\end{table}

\begin{table*}
\caption{Classification accuracy of multi-scale content features for both $BF$ and $FF$, and the aggregations (mean,max,min) on the Event8 dataset. 
The best result is in bold. }
\vspace{4mm}
\centering
\begin{tabular}{c|c|c|c|c|c|c|c|c|c}
\toprule
\label{tab:multiscale}
Content & 1.6$\times$& 1.4$\times$ & 1.2$\times$ & 1$\times$ & 0.8$\times$ & 0.6$\times$ & Mean & Max& Min\\
features & &  &  &  &  &  &  & & \\
\midrule
$FF$ &96.4&96.6  &96.8  & 97.0 &96.7  &96.8  &97.2 &\bf{97.2} &96.8 \\
\hline
$BF$ &93.1 & 93.7&94.2  &94.8  &95.4 &95.7 &95.3  &\bf{95.8} &93.8 \\
\bottomrule
\end{tabular}
\end{table*}
\subsection{Analysis of multi-scale content features}
\label{multiscale}

We analyze different scaled features individually as well as their aggregations (for multi-scale features) using pre-trained deep learning model, VGG16 \cite{simonyan2014very} for both content features based on both foreground and background information.
In particular, we study three different types of aggregation methods (mean, max and min) on the Event-8 dataset. The average classification accuracies are listed in Table \ref{tab:multiscale}. While observing the table, we notice that the best aggregation method for multi-scale features is the max pooling method. Thus, we select it as the aggregation method in this work.

\subsection{Statistical analysis of our method}
\label{statistical_analysis}
\begin{figure}
    \centering
    \includegraphics[width=0.90\textwidth, height=100mm,keepaspectratio]{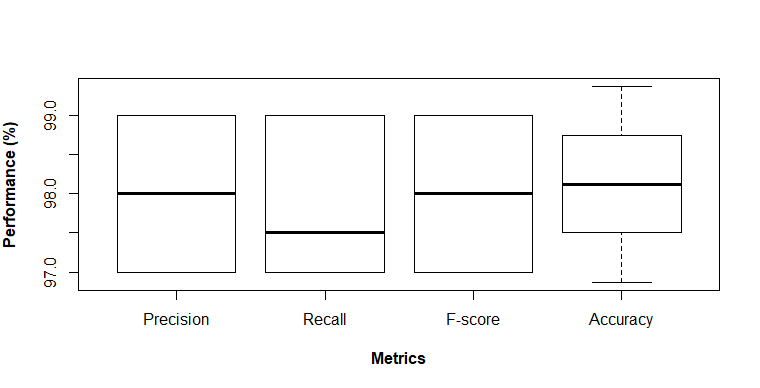}
    \caption{Boxplot of performance metrics achieved on Event-8 dataset.}
    \label{fig:boxplots}
\end{figure}

We analyze statistical analysis of our method ($CCF$) based on four performance metrics, Precision (Eq. \eqref{eq:precision}), Recall (Eq. \eqref{eq:recall}), F-score (Eq. \eqref{eq:fscore}), and Accuracy, using box-plot (refer to Fig. \ref{fig:boxplots}) on Event-8 dataset, which contains 10 runs of different training and testing sets. While observing Fig. \ref{fig:boxplots}, we notice that our method imparts promising performance on Event-8 for all four metrics (Precision, Recall, F-score, and Accuracy).
\begin{equation}
    \text{Precision} = \frac{TP}{TP+FP},
\label{eq:precision}
\end{equation}
\begin{equation}
  \text{Recall} = \frac{TP}{TP+FN},
 \label{eq:recall}   
\end{equation}
\begin{equation}
  \text{F-score} = 2\times \frac{(\text{Recall} \times \text{Precision})}{(\text{Recall} + \text{Precision})},
\label{eq:fscore}
\end{equation}
where $FP$, $TP$, $FN$ denote false positive, true positive, and false negative, respectively.

Similarly, we also analyze the significance test of our method ($CCF$) for each metrics using two-tailed t-test. We notice that 95\% confidence interval of Precision, Recall, F-score, and Accuracy are 98$\pm$  1.2 (p-value $\textless$ 2.2e-16), 97.80 $\pm$ 1.3 (p-value $\textless$ 2.2e-16), 98 $\pm$ 1.2 (p-value $\textless$ 2.2e-16), 98.1 $\pm$ 1.2 (p-value $\textless$ 2.2e-16), respectively.
To this end, through both statistical analysis and significance test, we conclude that our method imparts stable performance.

\section{Conclusion and Future Works}
\label{conclusion}

In this paper, we have proposed new methods to extract context and content features, and fuse them together to extract features of scene images. We exploit annotations/descriptions of similar images available on the internet for context features, and pre-trained deep learning models to extract background and foreground information for content features. The content features are multi-scale in nature, and the tag-based context features are based on tags related to the image.
Our experiments indicate that the designed features are suitable for scene image representation because of the ability in handling ambiguity due to intra-class dissimilarity and inter-class similarity.  

Experimental results on three commonly used benchmark datasets unveil that our proposed context and content features produce better classification results than many existing context and content features, respectively and the the fusion of proposed context and content features ($CCF$) produces significantly better results than numerous existing tag-based and deep features over all three datasets.

However, the content features (multi-scale content features) are not rotation-invariant. It may not work properly if the images are rotated. We would like to investigate this limitation and introduce rotation-invariant content features for higher discriminability. While extracting tag-based context features, it may not be possible to always have class labels. In this case, unsupervised learning on the context features is potential to detect some interesting discriminating patterns of tags. In the future, we would like to extend our work with resorting to unsupervised learning for the tag-based context features extraction.

\section*{Acknowledgement}
Chiranjibi Sitaula is supported by Deakin University Postgraduate Research Scholarship (DUPRS) award.
Dr Sunil Aryal is supported by a research grant funded jointly by the US Air Force Office of Scientific Research (AFOSR) and Office of Naval Research (ONR) Global under award number FA2386-20-1-4005.
Dr. Xuequan Lu is supported by Deakin internal grant (CY01-251301-F003-PJ03906-PG00447 and 251301-F003-PG00216-PJ03906).

\bibliographystyle{plain}
\bibliography{mybibfile}
\typeout{get arXiv to do 4 passes: Label(s) may have changed. Rerun}
\end{document}